\documentclass{ifacconf}

\usepackage{graphicx}
\usepackage{caption}
\usepackage{color}
\usepackage{natbib}    
\usepackage{kotex}
\usepackage{multirow}
\usepackage{adjustbox}
\usepackage{amsmath,amssymb,enumerate}

\usepackage{setspace}
\usepackage{booktabs}
\usepackage{mathtools}
\usepackage{mathrsfs}
\usepackage[mathscr]{euscript}
\usepackage{times}
\usepackage{cite} 
\usepackage{amsfonts}
\usepackage{textcomp}
\usepackage{arydshln}
\usepackage{balance}
\usepackage{subcaption}

\newcommand{\jwc}[1]{{\textcolor{black}{#1}}}

\begin{document}
\begin{frontmatter}

\title{Social Zone as a Barrier Function for Socially-Compliant Robot Navigation\thanksref{footnoteinfo}} 

\thanks[footnoteinfo]{This research was supported by NSF Award No. 2118818.}

\author[First]{Junwoo Jang} 
\author[First]{Maani Ghaffari} 

\address[First]{Department of Naval Architecture and Marine Engineering \\ University of Michigan, Ann Arbor, MI 48109, USA \\ e-mail: \texttt{\{junwoo, maanigj\}@umich.edu}}

\begin{abstract}        
This study addresses the challenge of integrating social norms into robot navigation, which is essential for ensuring that robots operate safely and efficiently in human-centric environments. Social norms, often unspoken and implicitly understood among people, are difficult to explicitly define and implement in robotic systems. To overcome this, we derive these norms from real human trajectory data, utilizing the comprehensive ATC dataset to identify the minimum social zones humans and robots must respect. These zones are integrated into the robot's navigation system by applying barrier functions, ensuring the robot consistently remains within the designated safety set. Simulation results demonstrate that our system effectively mimics human-like navigation strategies, such as passing on the right side and adjusting speed or pausing in constrained spaces. The proposed framework is versatile, easily comprehensible, and tunable, demonstrating the potential to advance the development of robots designed to navigate effectively in human-centric environments.
\end{abstract}


\end{frontmatter}

\section{Introduction}

Robots are designed as intelligent systems to assist humans by taking over dangerous or repetitive tasks. Socially assistive robots, for example, aid in household chores at home and guide visitors in large public spaces like museums and airports~\citep{fu2023human, kathuria2022providers}. These robots are becoming increasingly integrated into our daily lives, enhancing comfort and efficiency. As robots and humans coexist, it is essential for robots to inherently possess the capability to navigate toward their destinations while avoiding people and obstacles in human-centric spaces. 

Beyond simply avoiding physical collisions, robots must adhere to social norms by moving like humans. For example, when walking toward someone face-to-face, we naturally adjust our paths to avoid appearing as if we’re about to collide. Similarly, robots should know how to navigate human environments in a socially compliant and culturally aware manner. This capability is studied within a research field known as social navigation. Many studies on social navigation are underway, yet evaluating and employing them is challenging due to the diversity of scenarios~\citep{mavrogiannis2023core, francis2023principles}. There are many possible scenarios, such as navigating narrow paths, avoiding a group of people, or following a specific person, with diverse costs like collision safety, human comfort, robot politeness, and legibility. Therefore, algorithms and frameworks need to be developed in a form that easily accommodates these extensions.

Proximity is a classic and universally applicable factor that explains avoidance movements around people~\citep{svenstrup2010trajectory}. Humans naturally maintain certain distances from each other. Research on modeling various proxemics, or \textbf{social zones}, explores the personal space around individuals, which, if invaded, can cause discomfort~\citep{rios2015proxemics}. This field was initially defined as concentric circular zones around a person~\citep{hall1963system}, representing different levels of comfort. Later models introduced more complex shapes, such as egg-shaped zones emphasizing the importance of frontal space~\citep{hayduk1981shape, kirby2009companion} or asymmetrical with smaller spaces on the pedestrian's dominant side~\citep{wkas2006social, gerin2008characteristics}. Further studies have shown that personal spaces can be dynamic, depending on factors like speed or grouping~\citep{truong2016dynamic, neggers2022effect}, and people might have different social zones with robots~\citep{patompak2020learning}. \jwc{
Traditionally, these social zones have been identified in experimental environments where many factors are controlled and participants are aware of being observed.} However, the social zones can vary significantly depending on several aspects, including the surrounding environment, the density of people present, cultural norms, and regional differences.

To study natural human behavior, it is essential to investigate social zones based on data recorded from real human data. Recently, \citet{corbetta2018physics} and \citet{pouw2024high} investigated how people maintain distance and avoid each other by analyzing the trajectories of pedestrians in real life. However, when using natural human data, we have to consider that individuals may have different policies because they may perceive social zones differently, and sometimes they do not strictly respect others' social zones. With this in mind, we aim to learn the minimum social zone, which robots must always strictly adhere to.

Once social zones are learned from real-life data, we can develop socially compliant movement behaviors using the control-barrier function (CBF)~\citep{ames2019control}. We treat the social zone as a set of hard constraints that should not be violated, and CBF ensures that the robot always stays within this safe set. \jwc{CBF enables natural behaviors, such as slowing down to avoid collisions, and its myopic nature makes it less sensitive to uncertainties in the predicted future paths of pedestrians.} To handle dynamic pedestrians, we combine CBF with model predictive control (MPC)~\citep{teng2021toward,zeng2021safety}, which allows the system to account for future events within a given prediction horizon.

We demonstrate that our method can adhere to social norms across diverse scenarios. To the best of the author's knowledge, this is the first attempt to derive social zones from real-life data and apply this insight to robot control, enabling robots to exhibit behaviors that mimic human interactions.


\section{Learning social zone}

When robots move to avoid humans, it is important to ensure not only physical safety, which prevents collisions, but also psychological safety, which avoids causing disturbance or discomfort to people. People maintain a respectful distance from each other while passing by, preemptively taking actions to signal their intent not to intrude into personal space. Although the social zones formed through interactions in various situations are not defined by explicit rules, they are universally recognized and practiced. To quantify the social zone, research has been conducted where robots move at different speeds and angles, investigating the comfort levels perceived by people~\citep{neggers2022determining}. However, psychological studies within a laboratory setting may differ from actual human behavior, and there is an issue that these studies do not mimic all possible situations, nor do they accommodate the varying levels of comfort unique to each individual.

In this regard, we analyze the actual pedestrian trajectories to quantify the social zone. Pedestrian trajectory data has been crucial for prediction problems and is therefore publicly accessible~\citep{korbmacher2022review}. However, commonly used trajectory datasets such as ETH, UCY, and GC~\citep{pellegrini2009you, lerner2007crowds, robicquet2016learning, yi2015understanding} have recordings less than an hour, which makes it challenging to represent the variety of situations pedestrians encounter. On the other hand, the ATC dataset records the trajectories of people moving around a 900 $m^2$ shopping mall over 92 days, providing an extensive human trajectory dataset~\citep{brvsvcic2013person}. Consequently, we analyze the ATC dataset to derive the social zone.

As illustrated in Fig.1, the ATC dataset provides pedestrian trajectories across a broad area of the shopping mall. We specifically extracted data from the central square, considering only situations where two individuals encounter each other in a large open space unaffected by walls or other structures. Among the 92 days of recorded data, there were occasions when events held in the square hindered the availability of open space, and such instances were manually removed from the dataset. According to \citet{kitazawa2009pedestrian}, people avoid obstacles within a 1 $m$ by 4.5 $m$ range in their direction of gaze. \textcolor{black}{To obtain comprehensive trajectories where people encounter and avoid others from various angles, we defined a larger attentional space of 4 $m$ by 5 $m$.} We extracted data where: 1) the attentional space is contained within the rectangular central square area, 2) only one other pedestrian is present in this space for a duration of 3 seconds, 3) the other pedestrian is initially at least 1 $m$ away, and 4) the reference pedestrian is moving at a speed of at least 0.4 $m/s$. We collected trajectories of two individuals' interactions, either walking in the same direction or passing each other, as shown in Fig.2.

\begin{figure}[t]
\begin{center}
\includegraphics[width=8.2cm]{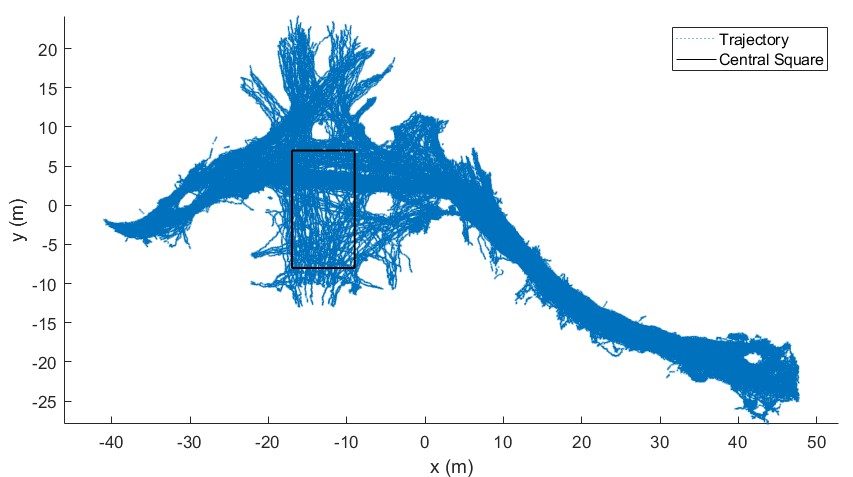}  
\caption{Pedestrian trajectories in the ATC shopping mall. The central square has a large open space and exhibits low pedestrian density, which is appropriate for investigating human interactions.} 
\label{fig:atc}
\end{center}
\end{figure}

\begin{figure}[t]
\begin{center}
\includegraphics[width=7.5cm]{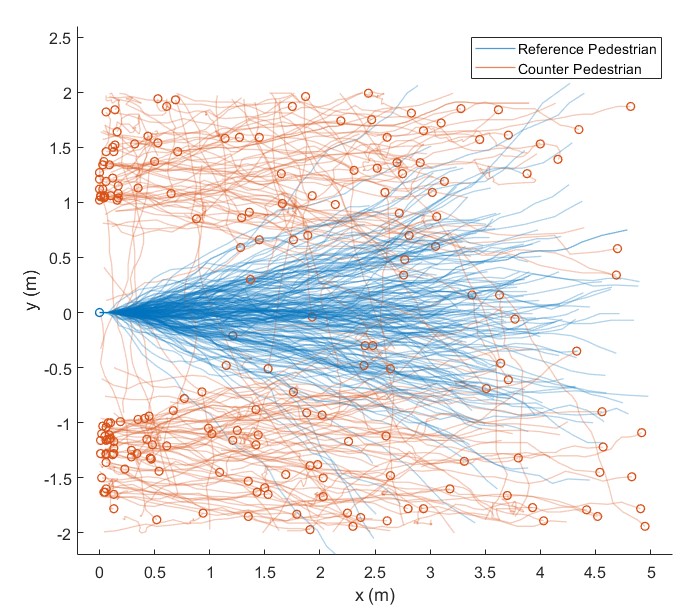}  
\caption{Examples of processed trajectories of two-person interactions from the open space. Since the data has been collected over an extended period, we can obtain trajectory scenarios of two individuals encountering each other from various speeds and directions. Although this shows 200 example trajectories, we have gathered a total of 16,181 trajectories.  } 
\label{fig:td}
\end{center}
\end{figure}

To derive social zones, we use the distance and line of sight (LOS) angle to other pedestrians at each moment. Figure 3 shows these distances according to LOS angles, providing rough information about minimum maintained distances between two pedestrians. Since the data is derived from the real world, outliers may occur, so we need to determine the minimum social zone that aligns with most situations. To remove outliers, we used the Local Outlier Factor (LOF)~\citep{breunig2000lof}, which operates based on local reachability density. Given the nature of LOF, where data density is low at the boundaries, it may be misclassified as an outlier. To prevent this, we defined the maximum distance in the data as $r_{\text{max}} = 2 \, m$, then calculated a complementary distance $r' = r_{\text{max}} - r$. The complementary distance according to angle can be represented in Cartesian coordinates, and we remove the outliers assuming an outlier fraction of 0.2 \%.

\begin{figure}[t]
\begin{center}
\begin{subfigure}[b]{0.5\textwidth}
     \centering
     \includegraphics[width=7.6cm]{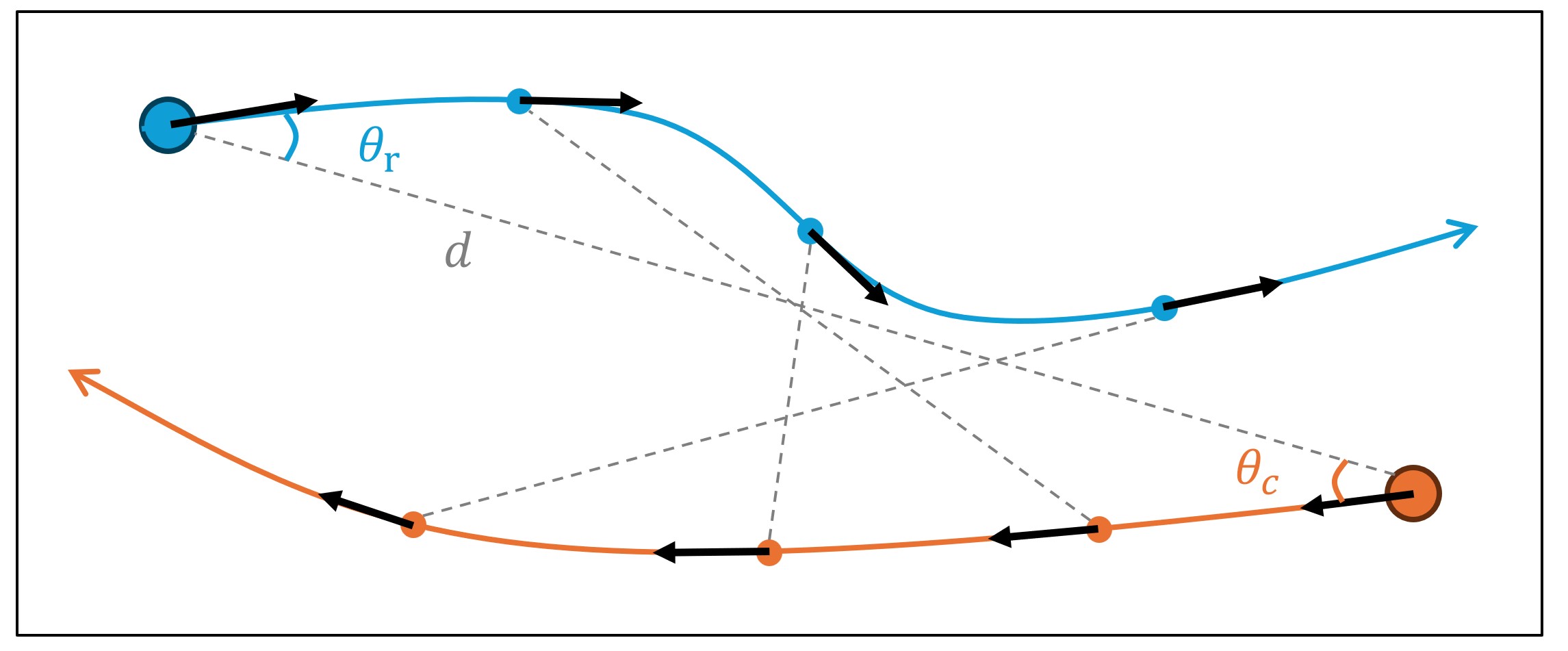}
     \caption{LOS angle and distance between each pedestrian}
     \label{fig:cd}
 \end{subfigure}
\\
\begin{subfigure}[b]{0.5\textwidth}
     \centering
     \includegraphics[width=8.2cm]{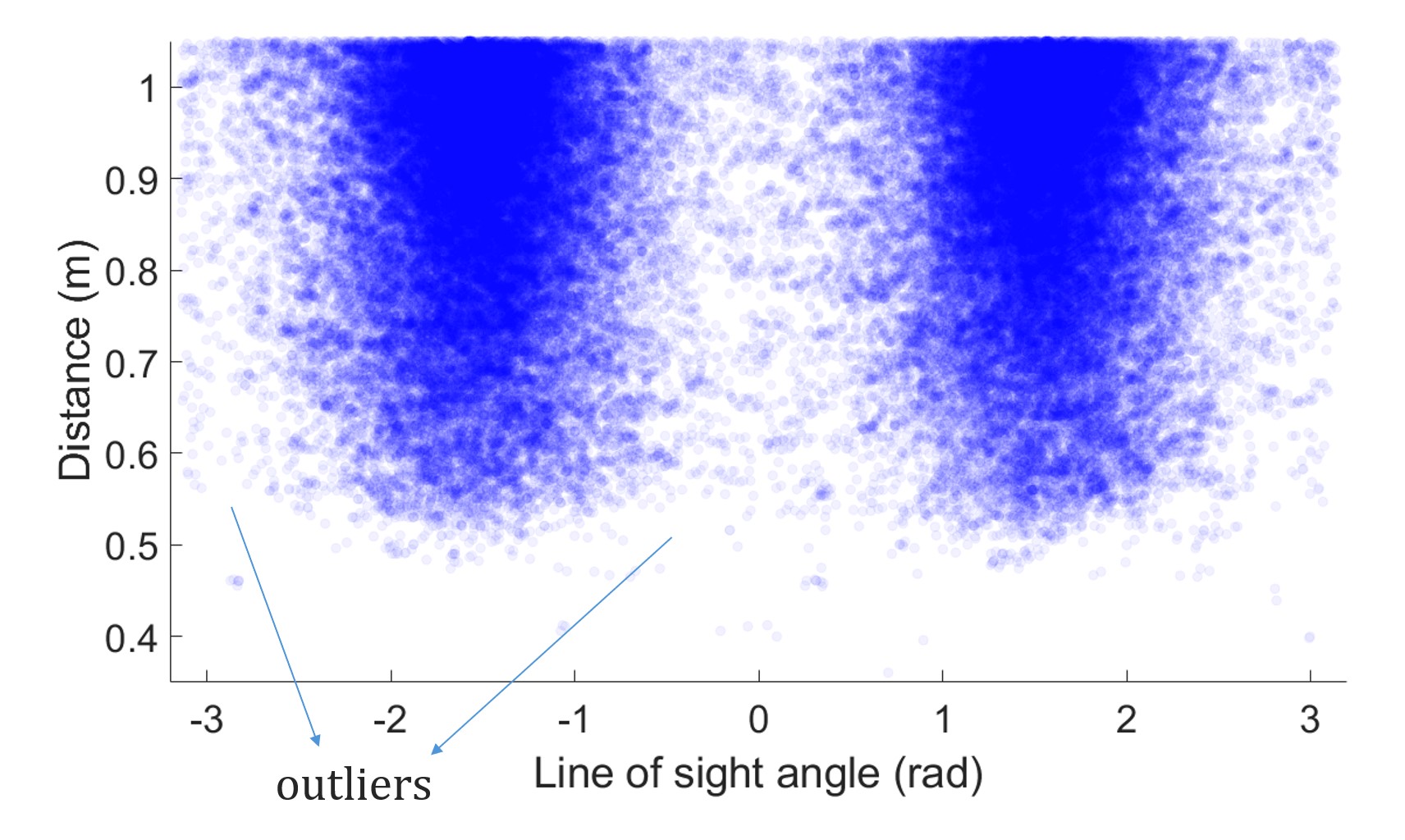}
     \caption{Correlation between LOS angle and distance}
     \label{fig:dd}
 \end{subfigure}
\caption{All distances based on the LOS angle derived from the discrete trajectories of two individuals. This roughly indicates the minimum social distance required for each angle of encounter. Given that this data comes from real-world observations, it may contain noise and outliers. Our goal is to establish the minimum boundary for the majority of the data.} 
\label{fig:df}
\end{center}
\end{figure}

\begin{figure}[t]
\begin{center}
\begin{subfigure}[b]{0.5\textwidth}
     \centering
     \includegraphics[width=7.6cm]{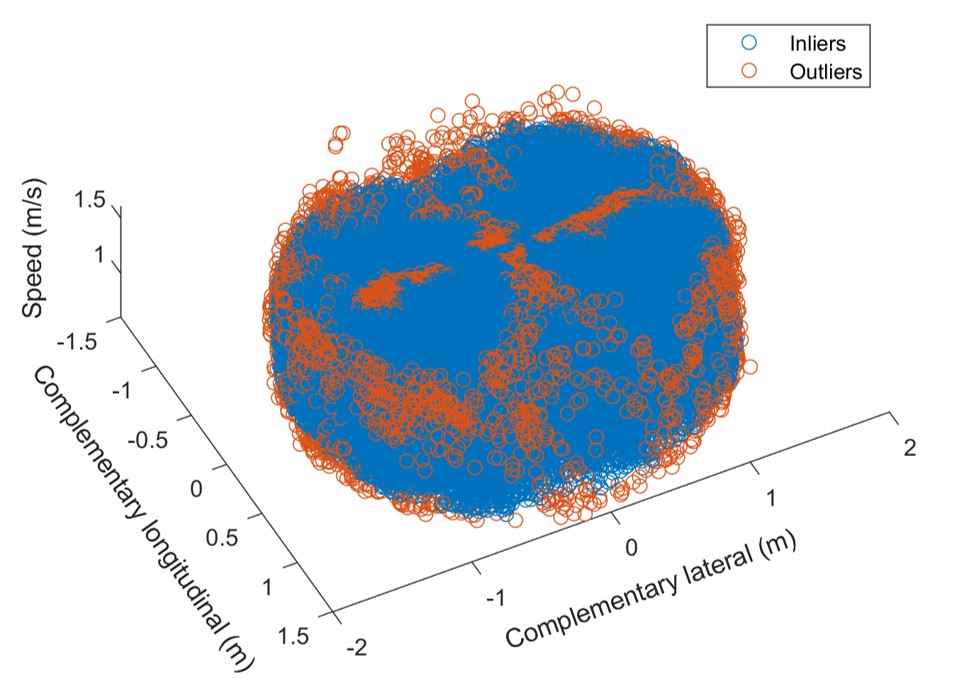}
     \caption{Inliers and outliers within the dataset.}
     \label{fig:ol}
 \end{subfigure}
\\
\begin{subfigure}[b]{0.5\textwidth}
     \centering
     \includegraphics[width=8.2cm]{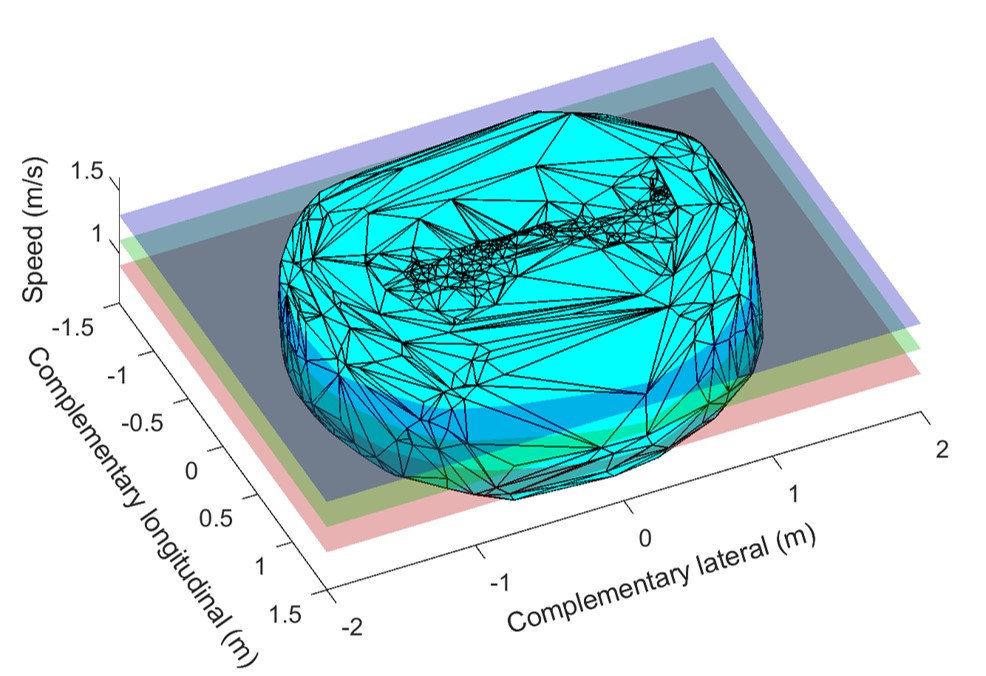}
     \caption{Convex hull enclosing the dataset and its intersections at different speeds.}
     \label{fig:is}
 \end{subfigure}
\caption{The distance data according to the reference pedestrian's speed, along with the dataset's outliers and enclosing convex hull. The distance dataset is represented through the complementary distance $r'$ to eliminate outliers effectively.} 
\label{fig:d}
\end{center}
\end{figure}

\begin{figure}[t]
\begin{center}
\includegraphics[width=7.4cm]{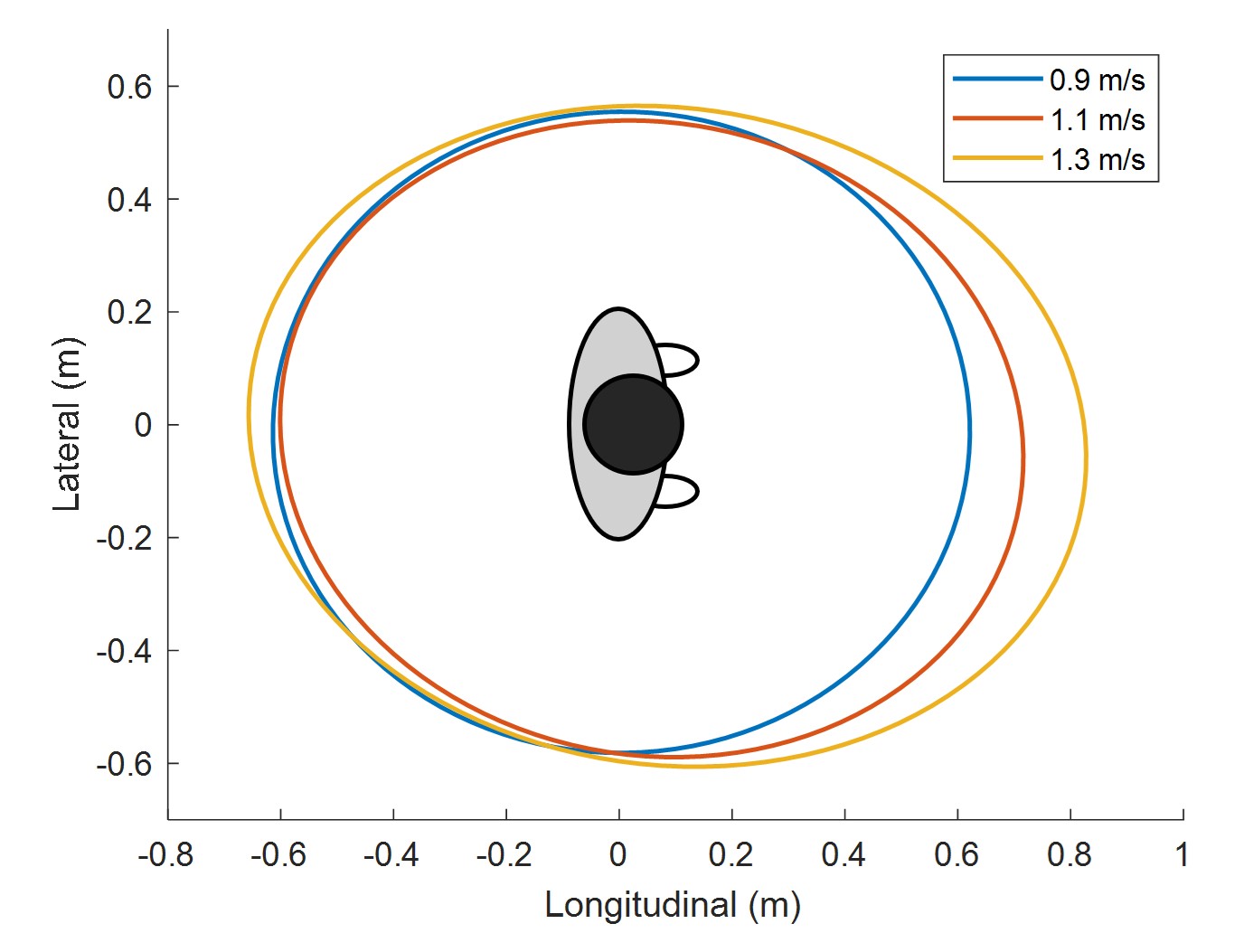}  
\caption{Estimated minimum social zones according to pedestrian speed from the ATC dataset.} 
\label{fig:sz}
\end{center}
\end{figure}

We can represent the data in 3D by adding the instantaneous speed of the reference pedestrian and determine the data's boundary by constructing a convex hull that encompasses all data points. The 3D convex hull generates a 2D polygon at the intersection with a plane defined by the speed axis, from which we can derive the minimum social zone according to the speed of the reference pedestrian. To simplify the representation of the minimum social zone depicted by the polygon, we have used minimum enclosing ellipse fitting to represent it as an ellipse~\citep{gartner1997smallest}. The obtained social zones are shown in Fig.~\ref{fig:sz}.

\section{Safety-guaranteed controller}
To control robots without invading the obtained minimum social zone, we utilize a CBF~\citep{ames2019control}. The safety set $S$ can be defined through a differentiable and continuous barrier function $h(\mathbf{x})$ as $S= \{\mathbf{x} \in \mathbb{R}^n : h(\mathbf{x}) \geq 0\}$. By designing a controller that ensures the barrier function remains positive, the system's trajectory can always reside within the safety set. In a discrete dynamic system model $ \mathbf{x}_{k+1} = f(\mathbf{x}_k, \mathbf{u}_k)$, a suitable barrier function can be achieved if there exists a class $\mathcal{K}$ function $\gamma$,
\begin{equation} \label{eq:gamma}
\Delta h(\mathbf{x}_k, \mathbf{u}_k) \geq -\gamma h(\mathbf{x}_k),
\end{equation}
where $\Delta h(\mathbf{x}_k, \mathbf{u}_k) := h(\mathbf{x}_{k+1}) - h(\mathbf{x}_k)$.

To simplify the problem, we use a scalar gamma where $0 < \gamma \leq 1$, and the lower bound of the control barrier function $h(\mathbf{x_k})$ diminishes exponentially at a rate of $1-\gamma$.


We address the problem of avoiding human social zones, which requires accounting for their dynamic behaviors in the CBF. Although the state of an obstacle can be defined in the system, we lack the exact dynamics model of humans and cannot control their movements.
Additionally, the increased complexity of the state and dynamics model can significantly increase the complexity of optimization. To address this, we can consider roughly predicting the movements of humans and integrating this with the MPC framework~\citep{zeng2021safety}. MPC optimizes control inputs iteratively based on model predictions, demonstrating robustness against noise or situation variations.
 
(MPC-CBF) Find $\mathbf{u}_k \in \mathcal{U}$ such that 
\begin{align}
\begin{split} \label{eq:mpc-cbf}
        \min_{\mathbf{u}_k}  \;\;\;\; &\mathbf{x}_N^\top P \mathbf{x}_N + \sum_{k=1}^{N-1} \mathbf{x}_k^\top Q \mathbf{x}_k + \mathbf{u}_k^\top R \mathbf{u}_k \\
        & s.t.  \;\;  \mathbf{x}_{k+1} = f(\mathbf{x}_k, \mathbf{u}_k)  \\
         & \mathbf{x}(0) = \mathbf{x}_0, \;\; x_k \in \mathcal{X}_k, \;\; \mathbf{u}_k \in \mathcal{U}_k, k = 0, 1, ..., N -1 \\
         & \Delta h(\mathbf{x}_k, \mathbf{u}_k) \geq -\gamma h(\mathbf{x}_k), \;\;  k = 0, 1, ..., N_h-1 
\end{split}
\end{align}
where $\mathcal{X} \subset \mathbb{R}^n$ and $\mathcal{U} \subset \mathbb{R}^m$ are the feasible state and control input sets, \textcolor{black}{$N$ and $N_h$ are the length of the prediction horizon for MPC and CBF constraints}, and $P$, $Q$, and $R$ are semi-positive definite cost matrices.

For the barrier function, it should satisfy $h(\mathbf{x}) = 0$ at the safe boundary and $h(\mathbf{x}) >0$ within other safe areas. Furthermore, an ideal barrier function would possess symmetry depending on the direction, ensuring it does not exhibit any particular bias toward being too evasive or close. In other words, when defined as a function influenced only by the distance from the boundary, $h(\mathbf{x}) = h(\mathbf{x}_r, \mathbf{x}_o) = h(d(\mathbf{x}_r, \mathbf{x}_o))$, where $d$ is the distance function between the states of the robot $\mathbf{x}_r$ and the obstacles $\mathbf{x}_o$, it enables consistent control to maintain distances regardless of the obstacle's shape. We assume that the general obstacle can be represented as line segments and use the following approximated distance function~\citep{shapiro1999implicit}. Let the endpoints of the line segment be $\mathbf{x_a} =(x_a, y_a)$ and $\mathbf{x_b} = (x_b, y_b)$ , the length $L=\|\mathbf{x_a} - \mathbf{x_b}\|$, and the midpoint  $\mathbf{x_c} = (\mathbf{x_a} + \mathbf{x_b})/2$. We define:
\begin{equation} \label{eq:g}
g(\mathbf{x}) := [(x-x_a)(y_b-y_a) - (y-y_a)(x_b-x_a)] /L,
\end{equation}
which is the signed distance function from point $\mathbf{x}$ to the line passing through $\mathbf{x_a}$ and $\mathbf{x_b}$.

A line segment can be represented as the intersection of an infinite line and a trimming region, such as a circular disk. We consider the following trimming function that is normalized to first order:
\begin{equation} \label{eq:t}
t(\mathbf{x}) = \frac{1}{L} [(L/2)^2 - \|\mathbf{x}-\mathbf{x_c}\|^2].
\end{equation}

With $g(\mathbf{x})$ and $t(\mathbf{x})$, a normalized distance function for the line segment,
\begin{equation} \label{eq:dist-line}
d(\mathbf{x}) = \sqrt{g(\mathbf{x})^2 + (\|t(\mathbf{x})\|-t(\mathbf{x}))^2/4},
\end{equation}
is zero exactly on the points of the line segment, positive everywhere else.

For an elliptical social zone, we define the distance function as a sum of distances from the two foci of the ellipse based on the fact that the sum of the distances from any point on the ellipse to two foci is constant. This provides an adequate approximation when the distance between the two foci of the ellipse is not too large. A general ellipse equation that is rotated by an angle $\theta$, centered at $(m, n)$ with $a$ and $b$ as the semi-major and semi-minor axes respectively, is given by:
\begin{equation} \label{eq:ellipse}
\begin{split}
&[(x-m)\cos(\theta) + (y-n)\sin(\theta)]^2/a^2 \\ &+ [(x-m)\sin(\theta) + (y-n)\cos(\theta)]^2/b^2 =1.    
\end{split}
\end{equation}
The foci are located at $\mathbf{c}_a = (m+c \cos(\theta), n+c \sin(\theta))$ and $\mathbf{c}_b = (m-c \cos(\theta), n -c \sin(\theta))$ where $c = \sqrt{a^2-b^2}$, and the distance function is defined as follows:
\begin{equation} \label{eq:dist-ellipse}
d(\mathbf{x}) = (\|\mathbf{x}-\mathbf{c}_a\| + \|\mathbf{x}-\mathbf{c}_b\|)/2 - a.
\end{equation}

\section{Simulation results}
In simulations, we demonstrate that the robot avoids a human using a minimum social zone and an MPC-CBF controller. We use a simple 2D double integrator model to describe the robot's dynamics:
\begin{equation} \label{eq:dd}
\mathbf{x}_{k+1} =
 \begin{bmatrix}
1 & 0 & \Delta t & 0 \\
0 & 1 & 0 & \Delta t \\
0 & 0 & 1 & 0 \\
0 & 0 & 0 & 1
\end{bmatrix}
\mathbf{x}_{k}+ \begin{bmatrix}
0.5 (\Delta t)^2 & 0 \\
0 & 0.5 (\Delta t)^2 \\
\Delta t & 0 \\
0 & \Delta t
\end{bmatrix}
\mathbf{u}_{k},
\end{equation}
where $\mathbf{x}_k =\{x,y,v_x, v_y\}$, $\mathbf{u}_k=\{f_x, f_y\}$, and $\Delta t$ is the time step.

It is assumed that the human moves at a constant speed of 0.5 $m/s$ and is unaffected by the robot's movements. The robot is assumed to have accurate knowledge of the human's position and velocity. The robot's maximum speed has been set at 1 $m/s$, and accordingly, it has been configured to always maintain a social zone of 1.1 $m/s$. The robot is modeled as a cylindrical shape with a radius \mbox{$r_r= 0.5$ $m$}, and a barrier function is designed to be greater than the robot's radius. However, since the approximated distance function contains some errors, an additional small margin has been set $\epsilon = 0.05$, $h(\mathbf{x}) = d(x)-r_r -\epsilon \geq 0$. The implementation of MPC-CBF utilizes the code from \citet{zeng2021safety}, and optimization was performed using the IPOPT \citep{wachter2006implementation} solver.

The simulation was conducted with a 0.1-second time step interval ($\Delta t$), with the MPC's prediction horizon ($N$) set at 8, and the CBF's horizon $(N_h)$ at 2. Despite having a shorter horizon than MPC, CBF can ensure stability, thereby increasing computational efficiency.

\begin{figure}[t]
\begin{center}
\begin{subfigure}[b]{0.5\textwidth}
     \centering
     \includegraphics[width=7.7cm]{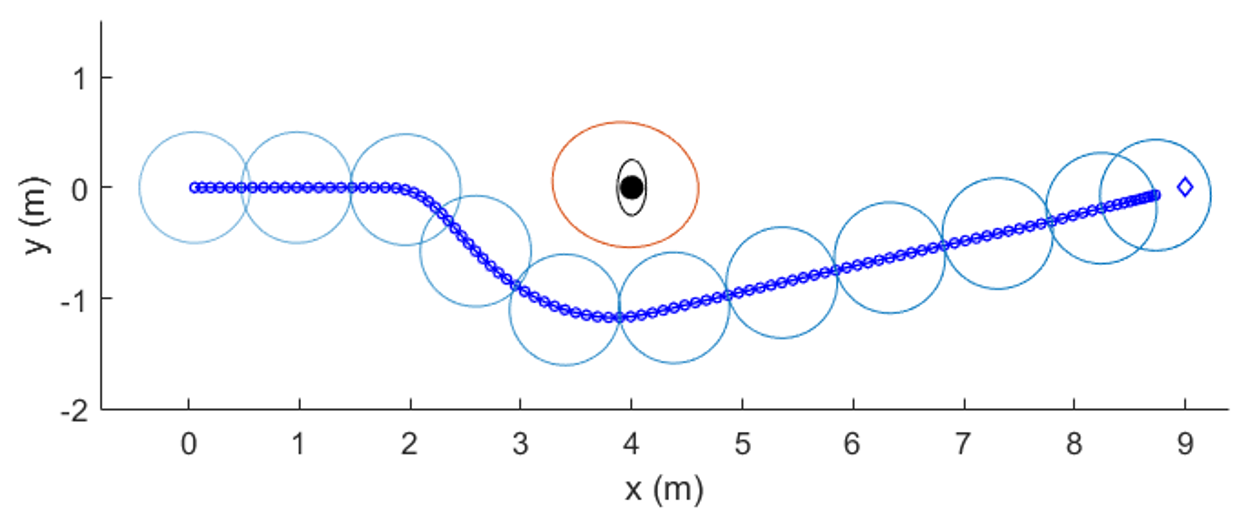}
     \caption{Passing a stationary person}
     \label{fig:sc1}
 \end{subfigure}
\\
\begin{subfigure}[b]{0.5\textwidth}
     \centering
     \includegraphics[width=7.7cm]{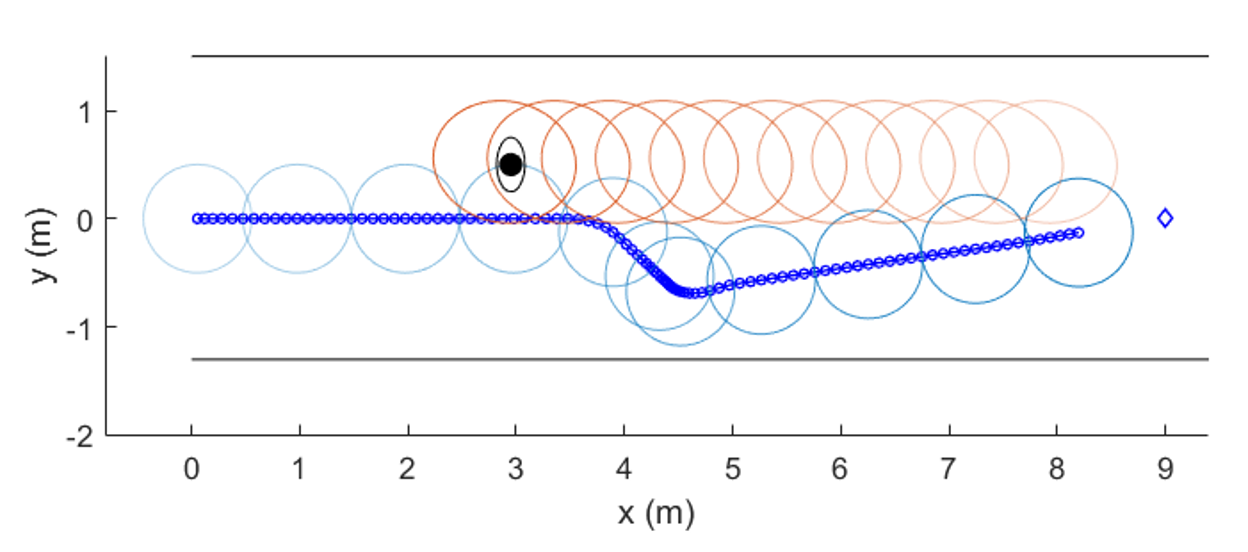}
     \caption{Avoiding a person in a narrow path}
     \label{fig:sc2}
 \end{subfigure}
 \\
 \begin{subfigure}[b]{0.5\textwidth}
     \centering
     \includegraphics[width=8.0cm]{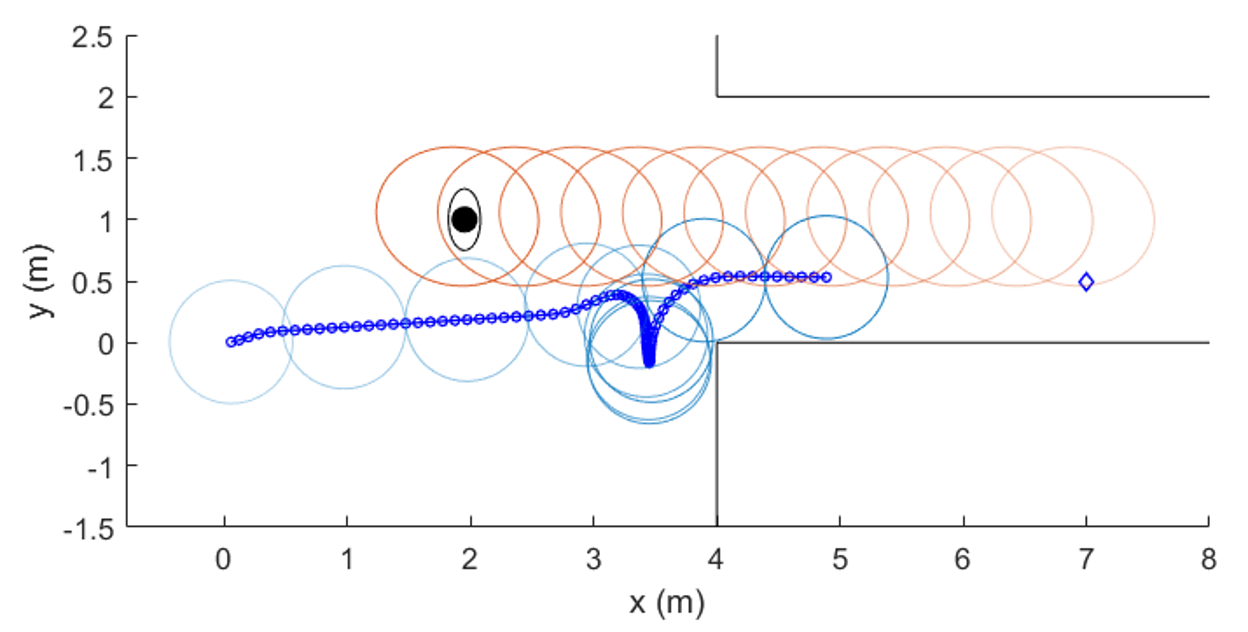}
     \caption{Yielding to a person in a restricted pathway (1) }
     \label{fig:sc3}
 \end{subfigure}
 \\
 \begin{subfigure}[b]{0.5\textwidth}
     \centering
     \includegraphics[width=8.0cm]{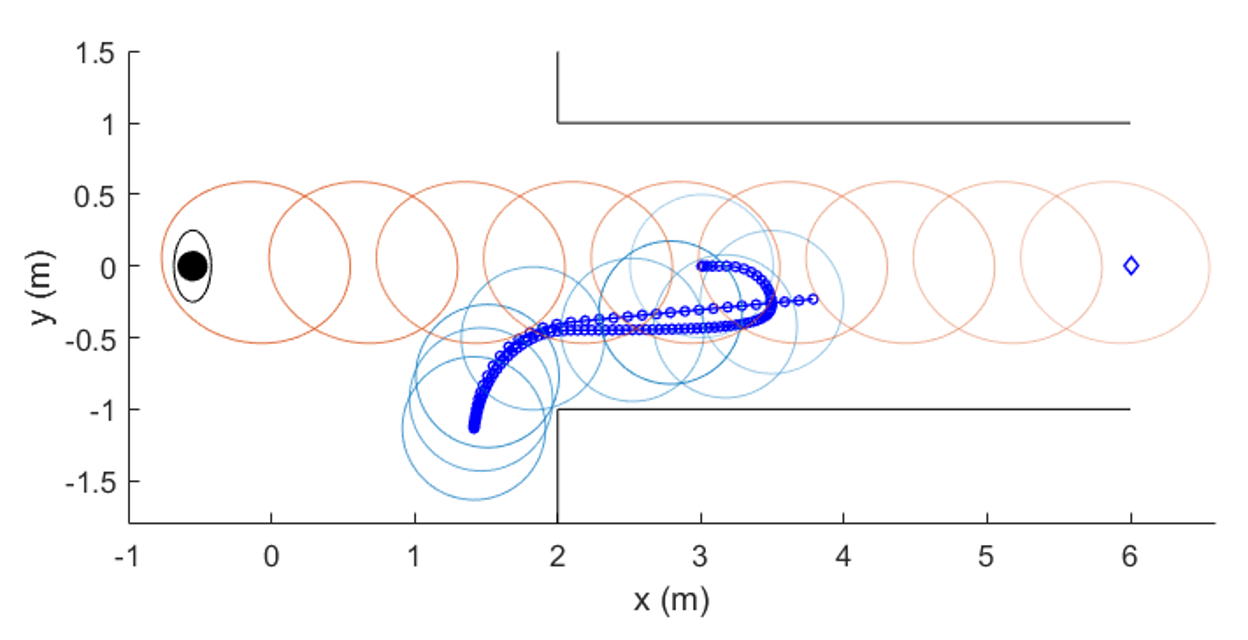}
     \caption{Yielding to a person in a restricted pathway (2)}
     \label{fig:sc4}
 \end{subfigure}
\caption{Robot and human interaction scenarios in various pathway conditions. 
The human's social zone is shown as a red ellipse, and the robot as a blue circle, with colors deepening over time to indicate progression. The robot's path is marked by circular markers. It targets a blue diamond, while a black line represents an obstructive wall.} 
\label{fig:sc}
\end{center}
\end{figure}

We investigate the robot's behavior in the following scenarios: 
\begin{enumerate}
\item Passing by a person facing directly;
\item Avoiding an approaching person in a narrow corridor;
\item Encountering a person in a restricted pathway.
\end{enumerate}
The third scenario shows two different cases depending on the positions of the robot and person. The results for these scenarios are depicted in Fig.~\ref{fig:sc}.

Each scenario demonstrates that the robot can appropriately follow social norms while avoiding humans. \jwc{Figure~\ref{fig:sc1} depicts a scenario where the robot navigates around a stationary person facing it. Because the social zone is larger in front of the person, the robot initiates its avoidance maneuver from a greater distance when approaching head-on. Although CBF is designed to prevent the robot from breaching the minimum social zone, this zone serves as a final boundary, and the actual avoidance maneuver begins well before reaching it. This early response signals a clear intent to avoid collision, like natural avoidance behaviors observed in real-world interactions.}

In Fig.~\ref{fig:sc2}, the robot is observed to reduce its speed when avoiding an approaching person in a narrow pathway. This is due to the CBF controller's feature. As the robot nears the safety boundary, the controller automatically slows the robot to enhance stability and prevent collisions. This cautious approach not only prevents collisions with walls and avoids intruding into the person's minimum social zone but also reassures humans of their safety by demonstrating that the robot will not cause harm, thereby providing a sense of comfort.

The results of scenario 3 in Fig.~\ref{fig:sc3} and~\ref{fig:sc4}, illustrate the robot's behavior strategies, to allow a person to pass by waiting or creating sufficient space before it proceeds. 
It is preferable for the robot to give priority to human movement. By yielding to humans, the robot not only empowers them to make decisions but also facilitates adaptable responses to unexpected situations, thereby enhancing safety. Moreover, this deference creates an environment where people can behave more naturally and autonomously, promoting a harmonious and effective integration of robotic systems into social settings. \textcolor{black}{The degree to which the robot yields can be adjusted by varying the size of the social zone, allowing us to appropriately trade off between the robot's yielding behavior and navigation efficiency.}


\section{Discussion}
\jwc{A major challenge in social robot navigation is explicitly defining objectives or cost functions. Recently, reinforcement learning and imitation learning have been employed to derive socially compliant behaviors from simulations or real human trajectories~\citep{kretzschmar2016socially, moller2021survey}. However, these approaches replicate human behavior policies without adequately explaining how they conform to social norms and struggle to generalize across different environments. In this study, we can better understand, fine-tune, and adapt the process to various scenarios by clearly defining the social zone and incorporating it into the control framework. This framework can be easily extended, for example, to different robot designs or varying the number of surrounding people. More importantly, using CBF ensures safety, offering a practical advantage over learning-based approaches.}

\jwc{As shown in the simulation results, integrating the social zone with CBF exhibits socially considerate behaviors, such as slowing down or yielding. Although a simple human motion prediction model was used in this study, the myopic design of the controller minimizes the impact of prediction errors on avoidance. However, these behaviors depend heavily on the choice of CBF’s barrier function and tuning parameters, as the defined minimum social zone only establishes a safety boundary at $h(\mathbf{x}) = 0$ and does not dictate behaviors outside this boundary. Future work could explore refining the barrier function for regions where $h(\mathbf{x}) > 0$ to more accurately replicate human-like avoidance strategies.}

\jwc{To further validate this methodology, we plan to test it in real environments. Due to differences between human and robot behavior policies, a larger social zone may be required, which can only be determined through real-world experiments involving robots. Because the degree of social compliance can be easily adjusted through CBF parameter tuning, incorporating feedback from participants during these experiments will allow us to iteratively refine the model, making the robot’s behavior more suitable for seamless integration into everyday life.}


\section{Conclusion}
We introduced a novel approach to socially compliant robot navigation by incorporating real-world human social zones into a robotic control system. Utilizing extensive real-life data, our method effectively addresses both physical and psychological aspects of human-robot interactions. Additionally, it extends to a navigation system that employs CBF and MPC to ensure safety amidst dynamic obstacles. Simulation results demonstrate that our approach enables the robot to adjust its behavior---like modulating speed, pausing, and yielding---showing strong potential for practical application. Moreover, identifying social zones deepens our understanding of space and human movement, which is crucial in human-centric environments.


{
\balance
\bibliography{ifacconf}       
}

                              
\end{document}